\documentclass{article}

\usepackage{microtype}
\usepackage{graphicx}
\usepackage{subfigure}
\usepackage{booktabs} 

\usepackage{bm}
\usepackage{tabularx}
\usepackage{multirow}

\usepackage{hyperref}

\usepackage[accepted]{icml2024}

\usepackage{amsmath}
\usepackage{amssymb}
\usepackage{mathtools}
\usepackage{amsthm}
\usepackage{gensymb}

\usepackage{hyperref}
\usepackage{balance}
\usepackage[capitalize,noabbrev]{cleveref}

\theoremstyle{plain}

\theoremstyle{definition}

\theoremstyle{remark}

\usepackage[textsize=tiny]{todonotes}

\icmltitlerunning{RAUCA: A Framework for Generating Robust and Accurate Camouflage}
\begin{document}

\twocolumn[
\icmltitle{RAUCA: A Novel Physical Adversarial Attack on Vehicle Detectors\\ via Robust and Accurate Camouflage Generation}




\begin{icmlauthorlist}
\icmlauthor{Jiawei Zhou}{sch1}
\icmlauthor{Linye Lyu}{sch1}
\icmlauthor{Daojing He}{sch1}
\icmlauthor{Yu Li}{sch1}
\end{icmlauthorlist}

\icmlaffiliation{sch1}{School of Computer Science and Technology, Harbin Institute of Technology, Shenzhen 518055, P.R. China}

\icmlcorrespondingauthor{Yu Li}{\texttt{li.yu@hit.edu.cn}}

\icmlkeywords{Machine Learning, ICML}

\vskip 0.3in
]



\printAffiliationsAndNotice{} 

\begin{abstract}

Adversarial camouflage is a widely used physical attack against vehicle detectors for its superiority in multi-view attack performance. One promising approach involves using differentiable neural renderers to facilitate adversarial camouflage optimization through gradient back-propagation. However, existing methods often struggle to capture environmental characteristics during the rendering process or produce adversarial textures that can precisely map to the target vehicle, resulting in suboptimal attack performance. Moreover, these approaches neglect diverse weather conditions, reducing the efficacy of generated camouflage across varying weather scenarios. To tackle these challenges, we propose a robust and accurate camouflage generation method, namely RAUCA. The core of RAUCA is a novel neural rendering component, Neural Renderer Plus (NRP), which can accurately project vehicle textures and render images with environmental characteristics such as lighting and weather. In addition, we integrate a multi-weather dataset for camouflage generation, leveraging the NRP to enhance the attack robustness. Experimental results on six popular object detectors show that RAUCA consistently outperforms existing methods in both simulation and real-world settings. 

\end{abstract}

\section{Introduction}
\label{submission}

\begin{figure}[t]
\vfill

\begin{center}
\centerline{\includegraphics[width=\columnwidth]{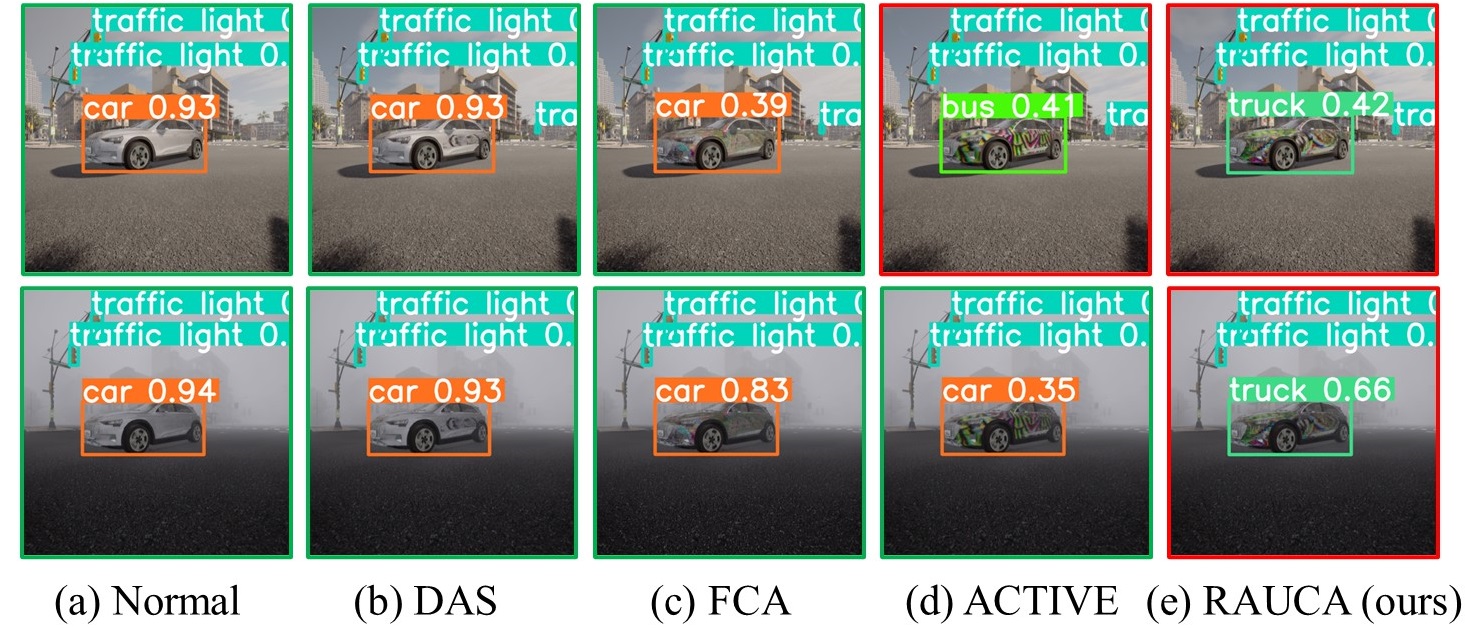}}
\vskip -0.1in
\caption{ Comparison of different adversarial camouflage under sunny (first row) and foggy (second row) environments, where only our method succeeds in both cases. (a) A car with normal texture; (b) DAS \cite{wang2021dual}. (c) and (d) are top-performed methods FCA \cite{wang2022fca} and ACTIVE \cite{Suryanto_2023_ICCV}, respectively. (e) Our method RAUCA. 
}

\label{introduction}
\end{center}
\vskip -0.45in
\vfill
\end{figure}

Deep Neural Networks (DNNs) have achieved remarkable performance in many real-world applications such as face recognition and autonomous vehicles \cite{krizhevsky2012imagenet,wang2023yolov7,chai2021deep}. However, DNNs suffer from adversarial examples \cite{szegedy2013intriguing}. For instance, in the task of vehicle detection,  adversarial inputs can deceive detection models, leading to incorrect detection of the surrounding vehicles, which poses a severe threat to the safety of autonomous vehicles.


Adversarial examples can be categorized into digital and physical adversarial types. Digital adversarial examples involve introducing pixel-level perturbations to the model's digital input, while physical ones manipulate real-world objects or surroundings, indirectly influencing the model inputs \cite{shapeshifter,Crime,Understanding}. 
Physical adversarial examples are generally deemed more practical, as gaining direct access to the model's input often necessitates system authentication. 
However, they are inherently more challenging as they must prove effective in complex physical environments, including various viewing perspectives, spatial distances, and lighting/weather conditions.

This paper focuses on physical adversarial examples against vehicle (e.g., car) detection models, given their wide adoption in autonomous driving scenarios where safety is of great importance. 
To ensure the attack effectiveness across various viewing angles, current methods prefer generating adversarial camouflage capable of covering the entire surface of the vehicle \cite{wang2021dual,wang2022fca,Suryanto_2022_CVPR,Suryanto_2023_ICCV}. 
Top-performing methods achieve this by leveraging a differential neural renderer. This renderer maps the 3D vehicle and its texture to 2D images, establishing a differentiable path between the 3D vehicle and the vehicle detection models. Thus, the texture can be optimized through gradient back-propagation for effective camouflage generation.

 In the literature, there are two ways to generate camouflage with a neural renderer: one way is to optimize a 2D square texture pattern and project it onto the vehicle repeatedly, referred to as world-align-based methods; the other is to optimize the 3D texture of the vehicle in the form of UV-maps, referred to as UV-map-based method. However, both approaches currently suffer from certain issues. The former method cannot guarantee the texture pattern projected onto the car in the same manner in both texture generation and evaluation, leading to differences in adversarial camouflage between generation and evaluation. The latter UV-map-based render methods are not capable of rendering sophisticated environment characteristics such as light and weather, leading to sub-optimal camouflage generation. Besides, all current methods fail to consider the effectiveness of the camouflage under various weather conditions. As shown in Figure \ref{introduction}, the camouflage generated by the top-performed FCA \cite{wang2022fca} and ACTIVE \cite{Suryanto_2023_ICCV} methods fails to attack the detector in a foggy scene.

To address the above issues, we develop a novel adversarial camouflage generation framework against vehicle detectors. Our key insight is that, for successful physical attacks, the generated camouflage must accurately map onto the vehicle. Additionally, the camouflage needs to be robust under different environmental conditions. Achieving the first goal necessitates an end-to-end optimization of the camouflage. The second goal requires a dataset encompassing ample environmental effects and also requires that these effects can be successfully utilized for camouflage optimization.
To fulfill these objectives, we leverage the UV-map-based neural renderer for accurate texture optimization. We adapt this renderer to achieve photo-realistic rendering of the target vehicle. Furthermore, we augment the dataset used for camouflage generation to include diverse weather conditions. This enhanced dataset, combined with the refined renderer, facilitates the generation of effective camouflage.



Our contributions can be summarized as follows:
\begin{itemize}
\item We present the Robust and Accurate UV-map-based Camouflage Attack (RAUCA), a framework for generating physical adversarial camouflage against vehicle detectors. It enhances the effectiveness and robustness of the adversarial camouflage through a novel rendering component and a multi-weather dataset.



\item We propose the Neural Renderer Plus (NRP), a novel neural rendering component that allows for the optimization of textures that can be accurately mapped to vehicle surfaces and can render images with environmental characteristics such as lighting and weather.


\item We incorporate a multi-weather dataset with ample environmental effects into the camouflage generation process. Our experiments show that the use of this dataset substantially enhances the attack robustness when using NRP for rendering.



\end{itemize}
Our extensive studies demonstrate that our method outperforms the current state-of-the-art method by around 10\% in car detection performance (e.g., AP@0.5). Our attacks achieve strong robustness and effectiveness, excelling in multi-weather, multi-view, and multi-distance conditions. We also demonstrate our method is effective in various weather conditions in the physical world. Our code is available at: \href{https://github.com/SeRAlab/Robust-and-Accurate-UV-map-based-Camouflage-Attack}{https://github.com/SeRAlab/Robust-and-Accurate-UV-map-based-Camouflage-Attack}.


 \section{Related work}
\textbf{Physical Adversarial Attack}: Physical adversarial attacks need to consider the robustness of the attacks because the objects and the environment are not constant. The Expectation over Transformation (EoT) \cite{athalye2018synthesizing} is a prime method of generating robust adversarial examples under various transformations, such as lighting conditions, viewing distances, angles, and background scenes. As a result, many adversarial camouflage methods \cite{zhang2019camou, wu2020physical,wang2021dual,wang2022fca,Suryanto_2022_CVPR,Suryanto_2023_ICCV} employ EoT-based algorithms to enhance their attack robustness in the real world scenarios.

\textbf{Adversarial Camouflage}:
For self-driving cars, precise detection of surrounding vehicles is a critical safety requirement. Consequently, there has been a growing interest in developing adversarial vehicle camouflage to evade vehicle detection systems. Current research works mostly leverage a 3D simulation environment to obtain 2D rendered vehicle images with various transformations such that they can obtain robust adversarial camouflage. The early researches of adversarial camouflage for vehicles are mostly black-box methods because the rendering process in the simulation environment is non-differentiable. \citet{zhang2019camou} first conducts experiments in the 3D space to generate adversarial camouflage for cars. They propose CAMOU, a method to train an approximate gradient network to mimic the behavior of both rendering and detection of the camouflage vehicles. Then, they can optimize the adversarial texture using this network. Meanwhile, \citet{wu2020physical} propose an adversarial camouflage generation framework based on a genetic algorithm to search an adversarial texture pattern. Then, they repeat the optimized texture pattern to build the 3D texture that covers the full-body vehicle. 

Recent methods introduce neural renderers, which enable differentiable rendering. With this technique, the adversarial texture can be optimized via gradient back-propagation. Two primary methods are employed in adversarial camouflage generation with neural renderer. One is to optimize a 3D model texture directly, which we refer to as UV-map-based camouflage methods; for example, \citet{wang2021dual} propose a Dual Attention Suppression (DAS) attack, which minimizes the model attention and human attention on the camouflaged vehicle. Besides, \citet{wang2022fca} propose the Full-coverage Camouflage Attack (FCA), which optimizes full-body surfaces of the vehicle in multi-view scenarios. The other is to optimize a 2D square texture pattern and then project it repeatedly to the target vehicle surface, which we refer to as world-align-based camouflage methods. \citet{Suryanto_2022_CVPR} present the Differentiable Transformer Attack (DTA), which proposes a differentiable renderer that can express physical and realistic characteristics (e.g., shadow). ACTIVE, proposed by \citet{Suryanto_2023_ICCV}, introduces a texture mapping technique that utilizes depth images and improves the naturalness of the camouflage by using larger texture resolution and background colors. 

While prior neural-renderer-based methods have achieved impressive attack success rates, they typically struggle to capture the environmental characteristics such as shadow or produce adversarial camouflage that can precisely map to the target vehicle, resulting in sub-optimal camouflage. Moreover, these methods frequently overlook diverse weather conditions during the generation of adversarial camouflage, hindering the effectiveness of the camouflage in varying weather environments.

\section{Method}

\begin{figure*}[t]
\vfill
\begin{center}
\centerline{\includegraphics[width=\columnwidth*2]{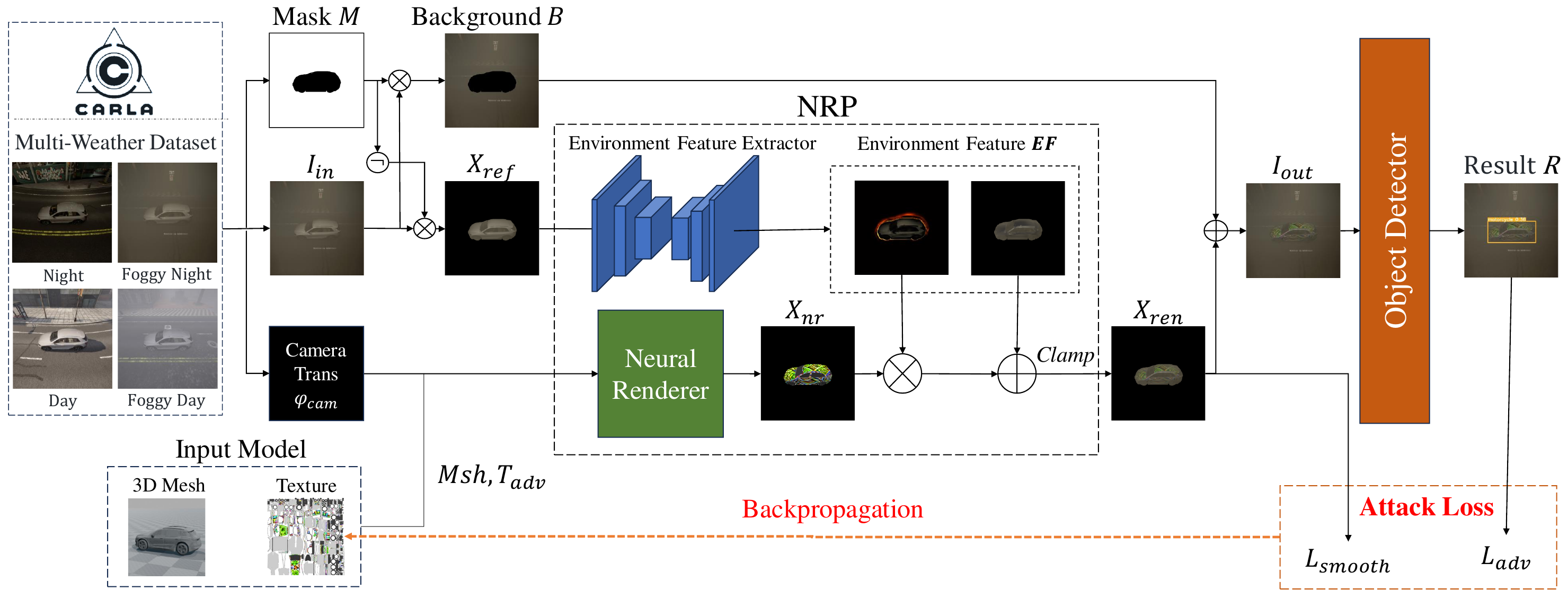}}
\vskip -0.1in
\caption{ The overview of RAUCA. First, a multi-weather dataset is created using CARLA, which includes car images, corresponding mask images, and camera angles. Then the car images are segmented using the mask images to obtain the foreground car and background images. The foreground car, together with the 3D model and the camera angle is passed through the NRP rendering component for rendering. The rendered image is then seamlessly integrated with the background. Finally, we optimize the adversarial camouflage through back-propagation with our devised loss function computed from the output of the object detector.}

\label{pipeline}
\end{center}
\vskip -0.4in
\vfill
\end{figure*}

In this section, we present the overview of our framework for generating adversarial camouflage. Subsequently, we provide an in-depth explanation of the essential components comprising our attack framework, RAUCA.

\subsection{Overview}

Figure \ref{pipeline} shows our entire framework for adversarial camouflage generation. Firstly, we modify the weather parameters in the simulation environment to
obtain a multi-weather vehicle dataset (\(I_{in}\), \(Y\), \(\Phi_{cam}\),
\(M\)), where \(I_{in}\) is the original  input images, \(Y\) is the
ground truth labels, \(\Phi_{cam}\) is the camera pose parameters
 (position and angle) for viewing the car, and \(M\) is the binary masks of \(I_{in}\), where the target vehicle areas are set to 0. With \(I_{in}\) and
\(M\), we can use:
\begin{gather}
X_{r e f}=I_{i n} \cdot \left (1-{M}\right)\label{1}\\
B=I_{in}\cdot {M}\label{2}
\end{gather}
to obtain the foreground car images \(X_{ref}\) and the ambient background images
\(B\). Then, we use the Neural Renderer Plus (NRP) \(N\), our proposal neural rendering component,
to obtain the rendered 2D vehicle images \(X_{ren}\ \) through
\begin{gather}
X_{r e n}=N\left (Msh, T_{adv}, \Phi_{c a m}, X_{r e f}\right)\text{,}\label{3}
\end{gather}
where \(Msh\) and \(T_{adv}\) are the 3D mesh and the UV texture map of the
vehicle, respectively. To obtain the realistic vehicle pictures, we apply a simple transformation
\(I_{out} = X_{ren} + B\) to attach the foreground \(X_{ren}\) to the corresponding
background. We then input \(I_{out}\) into the target detector \(D\) and
obtain the detection results  with \(R = D (I_{out})\).

Our framework aims to generate adversarial camouflages for vehicles to evade the detection of the vehicle detector. We can obtain the final adversarial camouflage through the solution of a specific optimization problem denoted as
\begin{gather}
\underset{T_{adv}} {\operatorname{argmax}} \mathcal{L}\left(D\left(N\left(Msh, T_{adv}, \Phi_{cam}, X_{ref}\right)+B\right), Y\right)\text{,}\label{4}
\end{gather}
where \(\mathcal{L}\) is our proposal loss
function.

\subsection{Multi-Weather Dataset}

According to Athalye et al.'s EOT study \cite{athalye2018synthesizing}, adding various weather conditions to training data notably boosts attack robustness. Nevertheless, real-world multi-weather dataset collection is hindered by high labor expenses, weather's inherent unpredictability, and difficulties in getting the mask \(M\) of the vehicle.

To address the above difficulties, we use CARLA \cite{dosovitskiy2017carla}, an autonomous driving simulation environment
based on Unreal Engine 4 (UE4), to obtain the multi-weather dataset. Modifying
the weather and time parameters with CARLA API to simulate different
weather and light environment conditions is convenient. Moreover, with its built-in semantic segmentation camera, we can accurately and conveniently segment foreground and background. 


 In the assembly of the Multi-Weather Dataset, as shown in Figure \ref{pipeline}, we strategically vary the sun altitude angle and fog density to simulate different weather conditions. The sun altitude angle is instrumental in modulating the intensity of sunlight within the environment, which in turn influences the light's interaction with the vehicle's camouflage. This variation in lighting can significantly affect the visibility and effectiveness of the camouflage. Concurrently, the density of fog, a critical environmental parameter, determines the extent to which the vehicle's surface is obscured. At specific densities, fog can effectively render parts of the texture partially or entirely invisible, a factor that is crucial in determining the success of adversarial textures.

\subsection{Neural Render Plus (NRP)}

\begin{figure}[t]
\vfill
\vskip 0.05in
\begin{center}
\centerline{\includegraphics[width=\columnwidth]{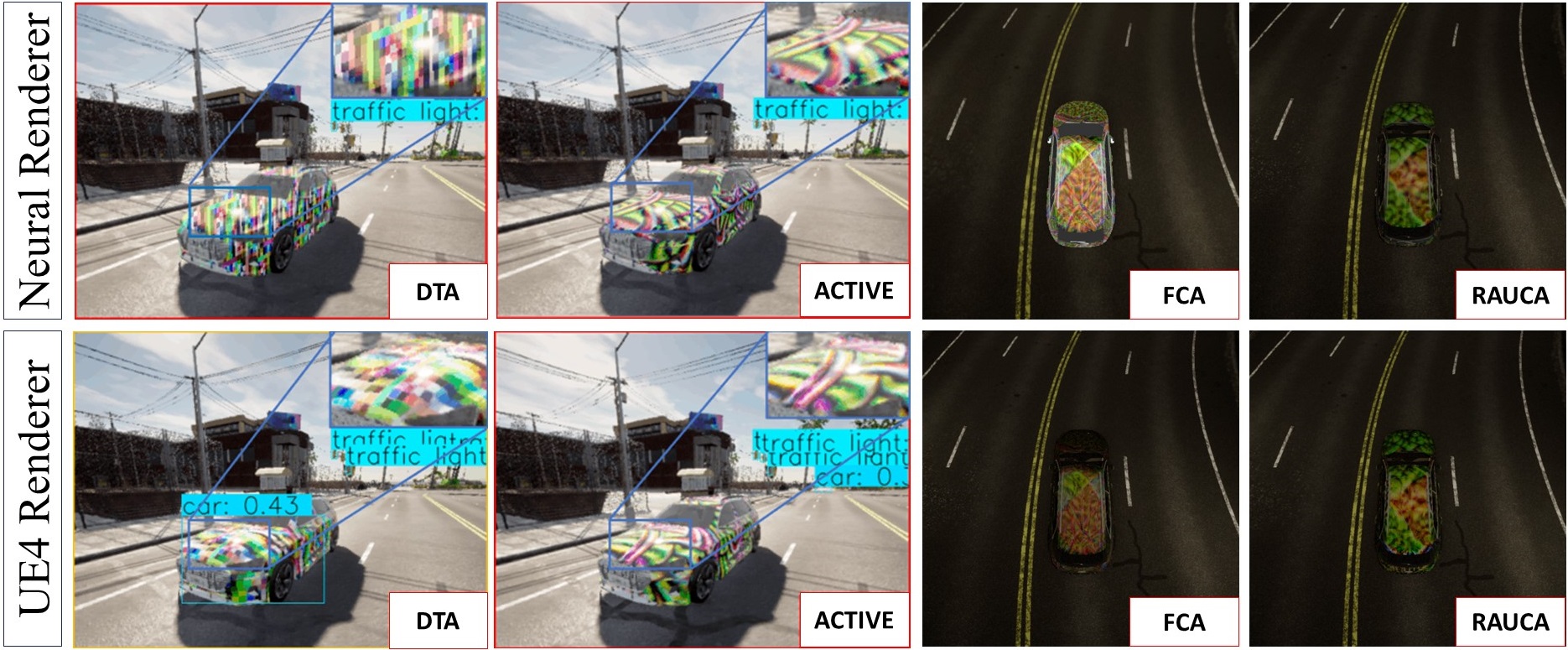}}
\vskip -0.05in
\caption{Comparison of rendering results of neural renderers used in different methods. The first row shows the results obtained by different neural renderers (already blended with the background) and the second row shows the rendered results in UE4. Our renderer is the only one that does both foreground environment rendering and texture rendering similar to UE4.}

\label{render}
\end{center}
\vskip -0.45in
\vfill
\end{figure}

In the rendering phase, we introduce NRP, a novel rendering
component that mitigates the limits of the previous neural-renderer-based camouflage methods. The world-align-based methods cannot accurately wrap the adversarial camouflage to the vehicle surface, which might weaken its attack performance in the real world. Hence, we choose a UV-map-based neural renderer to avoid this issue. However, the UV-map-based methods struggle to render the environmental characteristics on the vehicle surface, resulting in unrealistic rendered images. Therefore, we introduce the environment feature extractor that can combine the environmental characteristics and neural renderer output to obtain a realistic and accurate image of the camouflaged vehicle.  

Following \cite{wang2021dual,wang2022fca}, we use Neural Renderer (NR) \cite{kato2018neural} , whose input are 3D mesh \(Msh\), adversarial texture \(T_{adv}\) and camera angle \(\Phi_{cam}\), to generate a rendered image of camouflaged vehicle. However, NR uses its own light source during rendering, which makes it impossible to render complex environmental characteristics similar to UE4. To amend this, we adopt the method proposed in
DTA and ACTIVE \cite{Suryanto_2022_CVPR,Suryanto_2023_ICCV}, introducing an encoder-decoder network to extract the
environmental characteristics  \(X_{ref}\). Since the NR's output
\(X_{nr}\) already encompasses the shape, rotation, and texture transformation of the vehicle, our network only needs to learn the
transformation of environmental characteristics. We call this network as
Environment Feature Extractor (EFE). EFE's outputs are two maps of environment
features EF. We can fuse them with \(X_{nr}\) through pixel-by-pixel
multiplication and addition to get \(X_{ren}\), an image of a textured
vehicle with environmental characteristics. 

Before applying the framework to generate adversarial textures, NRP needs to be trained in advance to
optimize the parameters of EFE. The NRP training process inputs the masked images of the white car \(X_{ref}\), 3D mesh \(Msh\), poses of the camera \(\Phi_{cam}\), and multiple preset 3D texture T with different colors. At the same time, we need to obtain images of different vehicle colors based on \(\Phi_{cam}\) and the colors of T from the photo-realistic renderer. Then, we cut out the vehicle parts as the network output target \(TG\). 

For each iteration of the training process, we input $x_{ref} \in X_{ref}$, \(Msh\), \(T\) and $\varphi_{cam}\in \Phi_{cam}$ into NRP to get the rendering result \(x_{ren}\). 
To optimize the parameters of the network, we compute the following loss function
\begin{gather}
L_{EFE} (x_{ref})=W\left (x_{r e f}\right) B C E\left (x_{ren}, tg\right),\label{5}
\end{gather}
where BCE is the binary cross-entropy loss, and \(W (x_{ref})=\frac{h*w}{s}\) is a weight function. \(h\) and \(w\) are the height and width of
\(x_{ref}\) and \(s\) is the number of pixel points in the vehicle part of
\(x_{ref}\). We introduce \(W (x_{ref})\) to balance NRP's rendering optimization across various camera viewpoints. The original BCE loss, calculated over the entire image, unfairly prioritizes rendering when the car occupies a small area of the entire image. To address this, we multiply BCE by \(W (x_{ref})\) to compute the difference over the vehicle area, improving rendering performance for views where the car occupies a small image area.

Compared to NR, the rendering component used in FCA \cite{wang2022fca} and DAS \cite{wang2021dual}, our NRP considers the
environmental characteristics to make the final input image to the target detector more realistic. Compared to neural renderers used in DTA and ACTIVE \cite{Suryanto_2022_CVPR,Suryanto_2023_ICCV},
our rendering component can render the adversarial camouflage based on
UV mapping projection instead of world-aligned projection, which makes our textures more robust to multiple views. Moreover, our renderer has a more accurate projection of the texture compared to DTN and NSR, due to the use of NR for texture projection. 

As shown in Figure \ref{render}, the renderings obtained by DTA and ACTIVE\textquotesingle s renderers have a noticeable difference in the vehicle textures from that in UE4. Additionally, the FCA\textquotesingle s renderer is not able to represent the complexity of light and shadow information, showing a clear difference between the foreground and the background. In contrast, the result of our rendering component is relatively accurate both in terms of environmental characteristics and texture mapping.

\subsection{Attack Loss}
We propose a novel attack loss function that consists of two key components for improving attack effectiveness. The first component is the Intersection Over Union (IOU) between the object detection model's output detection box and the ground-truth box; the second component comprises the class confidence score and objectiveness score of the output. The loss function is denoted as
\begin{gather}
H_d (x)=\operatorname{IoU}\left (H_b\left (x\right), g t\right) * H_c\left (x\right) * H_o\left (x\right)\notag \\
L_{a t k}\left (x\right)=-\log \left (1-\max \left (H_d (x)\right)\right),\label{6}
\end{gather}
where \(x\) is the input image of the target detector, \(H_{b} (x)\) is
the detection bounding box, \(gt\) is the
ground-truth box and \(IoU\left ( H_{b} (x),gt \right)\) is the Intersection
over Union (IoU) between \(H_{b} (x)\) and \(gt\). We use
\(IoU\left ( H_{b} (x),gt \right)\) as a weight term, which allows the
optimized loss function to focus more on the bounding box with a larger
intersecting ratio area with \(gt\). \(H_{o} (x)\) and \(H_{c} (x)\)
represent the objectiveness score and the car class confidence score for the
bounding box, respectively. \(H_{d}\left. (x \right.)\) is our
detection score, which is the product of the objectiveness confidence,
class confidence, and the intersect ratio. Due to 
 \(IoU\left ( H_{b} (x),gt \right)\), we only assign a non-zero detection score to the detected boxes which are intersected with the ground-truth box. This makes the texture optimization focus on making the target vehicle detection ineffective. We select the highest \(H_{d} (x)\) and use it to compute \(L_{atk} (x)\) through a log loss. By minimizing \(L_{atk} (x)\), we can make the camouflaged vehicle misclassified or undetected by the object detector. 

\subsection{Smooth Loss}

To ensure the smoothness of the generated texture for human
vision, we follow \cite{sharif} to utilize smooth loss 
\(L_{sm}\) to enhance texture consistency. We use the output of NRP \(x_{ren}\) to compute for texture smoother in natural light and shadow. The smooth loss function can be defined as:
\begin{gather}
L_{sm}=\sum_{i, j}\left (x_{i, j}-x_{i+1, j}\right)^2+\left (x_{i, j}-x_{i, j+1}\right)^2,\label{7}
\end{gather}
where \(x_{i,j}\) is the pixel value of \(x_{ren}\) at coordinate (i, j).

Finally, our loss function \(L_{total}\) can be summarized as
\begin{gather}
L_{total }=\alpha L_{a t k}+\beta L_{sm},\label{8}
\end{gather}
where $\alpha$\label{alpha}, $\beta$\label{beta}  are the hyperparameters to control the contribution of each
loss function.

\section{Experiments}

In this section, we first describe our experimental settings. Then we
evaluate the attack effectiveness
of our adversarial camouflage in both simulated
and physical environments.

\subsection{Experimental Settings}

\textbf{Datasets}: We utilize CARLA to generate datasets in our experimentation. To facilitate a comparative analysis with previous studies \cite{wang2021dual,wang2022fca,Suryanto_2022_CVPR,Suryanto_2023_ICCV}, we select the Audi E-Tron as the target vehicle model. We create datasets using various simulation settings. For NRP training and testing, we utilized 69,120 and 59,152 photo-realistic images encompassing 16 distinct weather conditions, respectively. These conditions are a combination of four sun altitudes and four fog densities. Additionally, we generate a set of 40,960 images for our texture generation and prepare 7593 images for each adversarial camouflage in robust evaluation. The weather conditions present in these two datasets are the same conditions employed during the training and testing of NRP. For transferability evaluation, we utilize an ``unseen-weather dataset" consisting of 7593 images in 16 novel weather conditions that are not seen during the texture generation phase. Moreover, we conduct experiments in the real world by printing five types of adversarial camouflages and sticking them to the 1:12 Audi E-Tron car models. We shoot 432 photos under different light conditions and 120 photos in a foggy environment for each model. Additional information regarding the construction process of the datasets can be found in the appendix.


\textbf{Compared methods}: We compare our framework with the state-of-the-art adversarial camouflage methods: DAS \cite{wang2021dual}, FCA \cite{wang2022fca}, DTA \cite{Suryanto_2022_CVPR}, and ACTIVE \cite{Suryanto_2023_ICCV}. Both DAS and FCA are UV-map-based methods. They optimize the 3D texture by minimizing the attention-map-based scores and the detection scores of the detection model, respectively. Meanwhile, DTA and ACTIVE are both world-align-based methods. They optimize a square texture pattern with a neural network that can project the texture onto the target object, whereas ACTIVE uses a projection network that is closer to the world-aligned projection in UE4. We compare our results using the official carefully optimized textures generated by these methods. Given the similarity in experimental setups between our approach and DAS/FCA, as well as the generality of the textures generated by DTA and ACTIVE, we consider that the comparison is fair.


\textbf{Evaluation metrics}\label{evaluation-metrics}: To evaluate the NRP rendering component, Mean Absolute Error (MAE) is computed to quantify the difference between the output of NRP and the established ground truth, particularly within the vehicle region. Furthermore, we evaluate the attack effectiveness of the adversarial camouflage with the AP@0.5 \cite{everingham2015pascal}, a standard benchmark reflecting both the recall and precision value when the detection IOU threshold is 0.5.

\textbf{Target detection models}\label{target-models}: Aligning with previous studies, we adopt YOLOv3 \cite{redmon2018YOLOv3} as the white-box target detection model for adversarial camouflage generation. To evaluate the effectiveness of the optimized camouflage, we utilize a suite of widely used object detection models, treated as black-box models except for YOLOv3. This suite includes YOLOX \cite{ge2021yolox}, Deformable DETR (DDTR) \cite{zhu2020deformable}, Dynamic R-CNN (DRCN) \cite{zhang2020dynamic}, Sparse R-CNN (SRCN) \cite{sun2021sparse}, and Faster R-CNN (FrRCNN) \cite{faster-rcnn}, all of which are pretrained on the COCO dataset and implemented in MMDetection \cite{chen2019mmdetection}.

\textbf{Training details}\label{implementation-details}: We utilize the Adam optimizer with a learning rate of 0.01 for NRP training and texture generation. We train the NRP over a span of 20 epochs and select the model exhibiting the best performance on the testing dataset. Aligning with Wang's study \yrcite{wang2022fca}, we configure the $\alpha$ and $\beta$ (See Eq. \ref{8}) values at 1 and 0.0001, respectively. We directly obtain the mask of the vehicle from the semantic segmentation camera feature in CARLA \cite{dosovitskiy2017carla}. During the adversarial camouflage generation phase, the camouflage texture is initialized randomly and trains with five epochs. We conduct experiments on a cluster with four NVIDIA RTX 3090 24GB GPUs.

\subsection{Evaluation in Physically-Based Simulation Settings}

In this section, we conduct a comparative analysis of RAUCA against current advanced adversarial camouflage attack methods, including DAS, FCA, DTA, and ACTIVE.

\begin{table}[t]
\vskip -0.1in
\caption{Comparison of the effectiveness of camouflages across various object detection models. Values are AP@0.5 (\%) of the car.}
\label{sim-table2}
\vskip 0.1in
\begin{center}
\begin{small}
\begin{sc}
\resizebox{1\columnwidth}{!}{
\begin{tabular}{lcccccc}
\toprule
\multirow{2}*{\textbf{Methods}} & \multicolumn{3}{c}{\textbf{Single-stage}} & \multicolumn{3}{c}{\textbf{Two-stage}} \\ 
\cmidrule(lr){2-4}\cmidrule(lr){5-7}
                         & YOLOv3            & YOLOX     & DDTR       & DRCN          & SRCN           & FrRCN                   \\ 
\midrule
Normal                   & 0.70              & 0.899         & 0.780     & 0.784             & 0.785     & 0.761             \\ 
Random                   & 0.623              & 0.812       &   0.581     & 0.670             & 0.659     & 0.652             \\ 
DAS                      & 0.638              & 0.875       &   0.707     & 0.709             & 0.710      & 0.710             \\ 
FCA                      & 0.555              & 0.795        &   0.555    & 0.655             & 0.619        & 0.654             \\ 
DTA                      & 0.507          &    0.692     &  0.341     & 0.592          & 0.461       & 0.534            \\ 
ACTIVE                   & 0.439           &    0.625      &  0.384    & 0.513          & 0.464       & 0.496             \\ \midrule
RAUCA             & $\bm{0.304}$       & $\bm{0.611}$   & $\bm{0.285}$     & $\bm{0.449}$        & $\bm{0.384}$     & $\bm{0.406}$             \\
\bottomrule
\end{tabular}
}
\end{sc}
\end{small}
\end{center}
\vskip -0.3in
\end{table}

\begin{figure*}[ht]
\begin{minipage}[t]{0.48\linewidth}
\centering
\includegraphics[width=\columnwidth]{{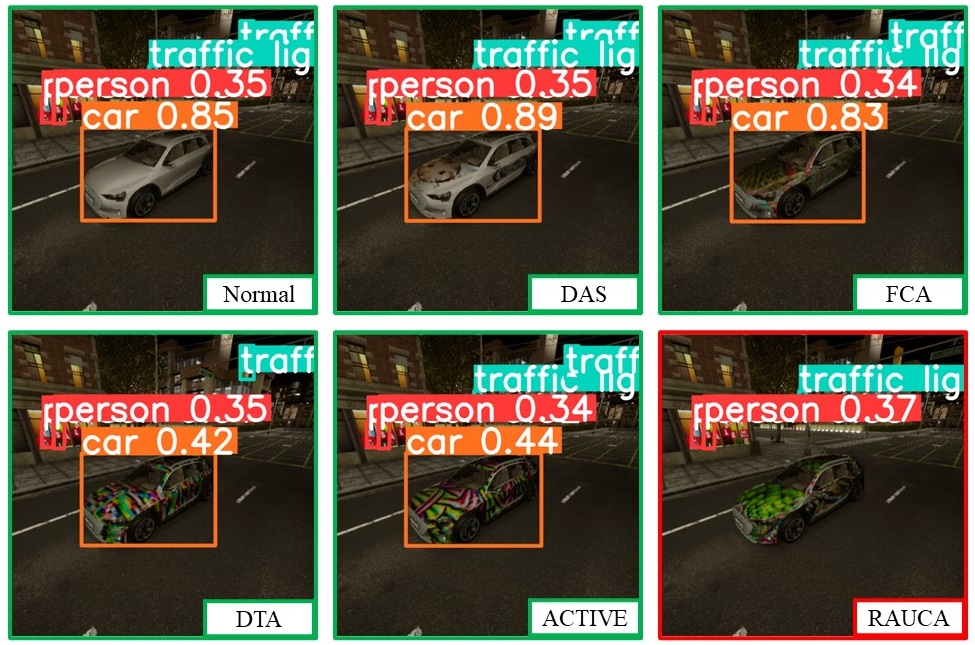}}
\vskip -0.13in
\caption{Attack comparison at night, a weather condition that has been included in our training set.}
\label{exp1}
\end{minipage}%
\hspace{0.6cm} 
\begin{minipage}[t]{0.48\linewidth}
\centering
\includegraphics[width=\columnwidth]{{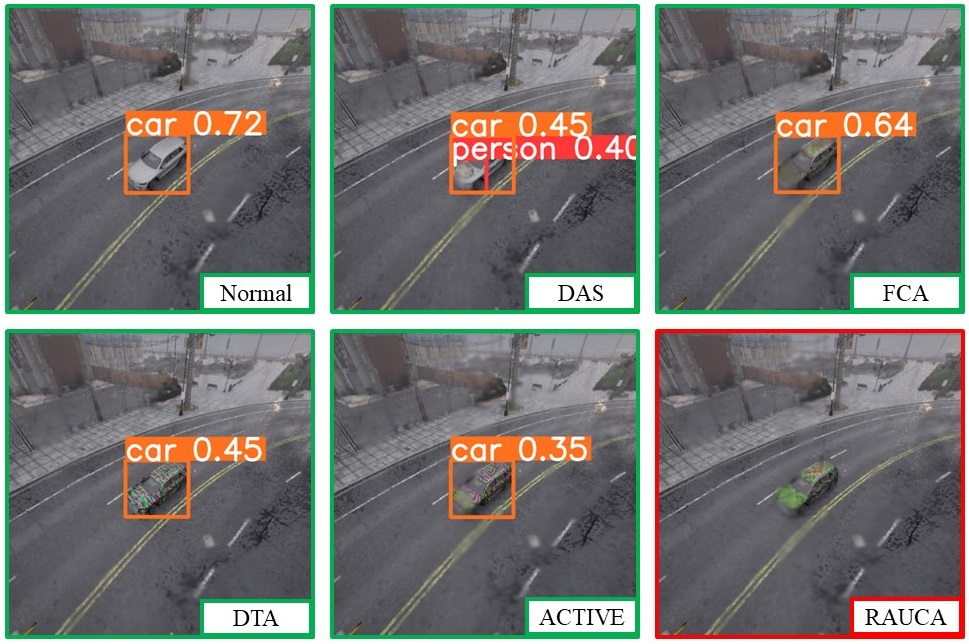}}
\vskip -0.14in
\caption{Attack comparison in the rainy day, a weather condition that hasn't appeared in our training set.}
\label{exp2}
\end{minipage}
\vskip -0.2in
\end{figure*}


\begin{figure}[t]
\vfill
\begin{center}
\centerline{\includegraphics[width=\columnwidth]{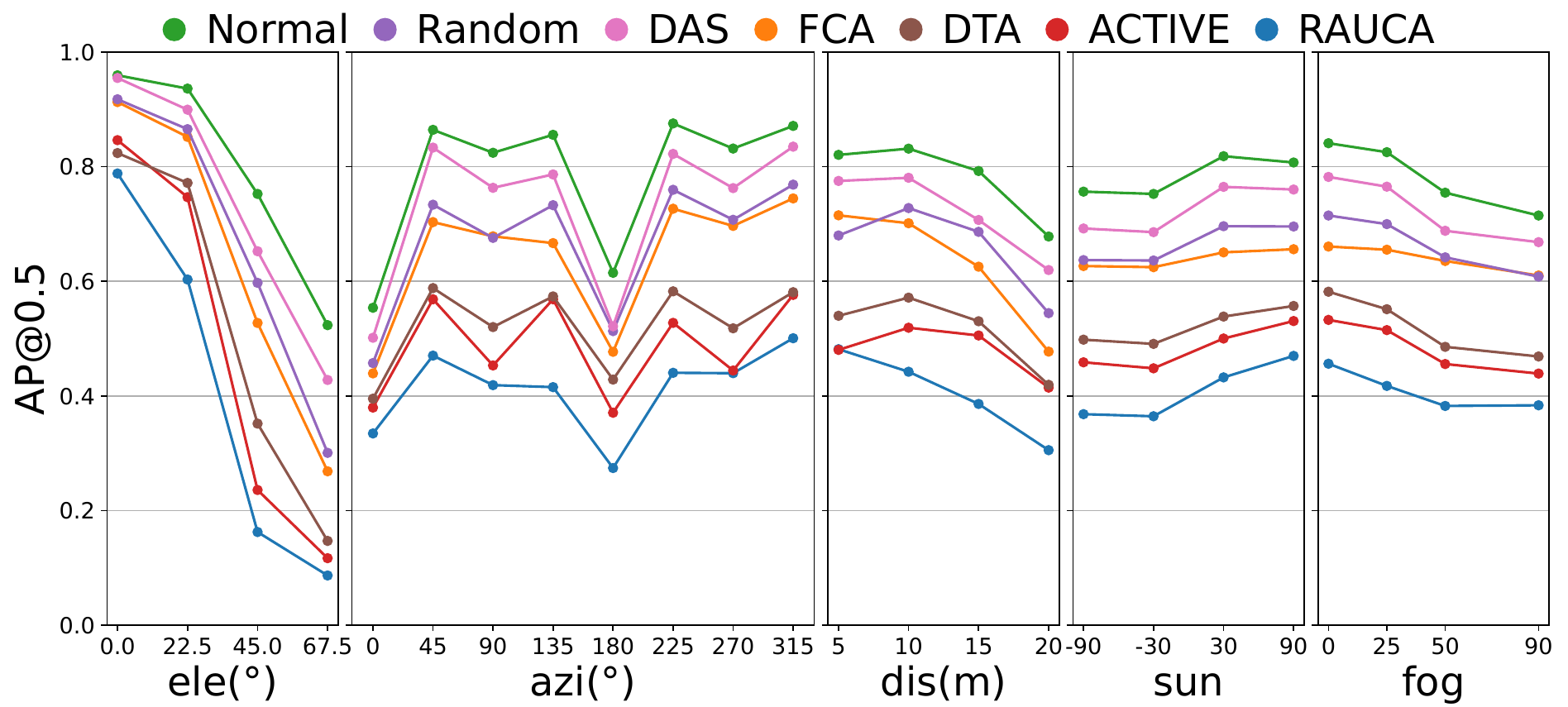}}
\vskip -0.15in
\caption{Attack comparison on different camera poses and weather parameters. ``ele" denotes elevation, ``azi" denotes azimuth, ``dis" denotes distance, ``fog" denotes fog density, and ``sun" denotes sun altitude angle. Values are car AP@0.5 (\%) averaged from all models.}
\label{curve}
\end{center}
\vskip -0.35in
\vfill
\end{figure}

\textbf{Robustness in multi-weather dataset}\label{4.2.1}: We run an extensive attack comparison using diverse detection models. In addition to YOLOv3, we also use various black-box detection models to evaluate the camouflage.
The results are shown in Table \ref{sim-table2}, showing that RAUCA has the best performance on all models. DAS generally performs less effectively, primarily due to the limitations of partially painted camouflage. Meanwhile, FCA exhibits sub-optimal performance, only slightly better than random camouflage, because it is not capable of rendering sophisticated environment characteristics during texture generation. DTA and ACTIVE are also less effective than our method because they do not consider multi-weather conditions during texture generation. Besides, their texture projection is not consistent between generation and testing. Our method demonstrates a nearly 13\% improvement in attack effectiveness compared to other models on the white-box YOLOv3 model. Additionally, it achieves an improvement of around 8\% on most other black-box models.
Figure \ref{exp1} shows an example of different vehicle camouflage at night: our method is still effective, while other methods fail to attack. 


Figure \ref{curve} shows the summarized performance of each camera pose and
weather parameter; values are car AP@0.5 averaged from
the detectors used in Table \ref{sim-table2}. We can see that the texture produced by our method outperforms other approaches in most
camera transformations and weather conditions. Our method improves multi-view robustness over previous methods in most viewpoints, thanks to the UV mapping-based projection for our textures and effective rendering of the environmental characteristics in the foreground. In addition, due to the incorporation of a multi-weather dataset during camouflage generation, our method is also the most robust in different weather environments.

\begin{table}[t]
\vskip 0.03in
\caption{Attack comparison in unseen-weather dataset. Values are the car AP@0.5 (\%).}
\vskip 0.07in
\label{sim-table3}
\begin{center}
\begin{small}
\begin{sc}
\resizebox{1\columnwidth}{!}{
\begin{tabular}{lcccccc}
\toprule
\multirow{2}*{\textbf{Methods}} & \multicolumn{3}{c}{\textbf{Single-stage}}& \multicolumn{3}{c}{\textbf{Two-stage}}  \\ 
\cmidrule(lr){2-4}\cmidrule(lr){5-7}
                         & YOLOv3            & YOLOX      & DDTR      & DRCN          & SRCN                     & FrRCN          \\ 
\midrule
Normal                   & 0.814              & 0.923         & 0.815      & 0.816             & 0.807                & 0.813             \\ 
Random                   & 0.752              & 0.837          & 0.597     & 0.698             & 0.651               & 0.691             \\ 
DAS                      & 0.757              & 0.887          & 0.739      & 0.730             & 0.725                  & 0.744             \\ 
FCA                      & 0.634              & 0.808           & 0.533     & 0.641             & 0.576                & 0.655             \\ 
DTA                      & 0.651              & 0.711          & 0.328      & 0.629             & 0.458                 & 0.545             \\ 
ACTIVE                   & 0.594              & 0.643         & 0.372      & 0.533             & 0.446                  & 0.513             \\ \midrule
RAUCA                     & $\bm{0.369}$              & $\bm{0.608}$    & $\bm{0.272}$              & $\bm{0.426}$             & $\bm{0.349}$                         & $\bm{0.395}$             \\ 
\bottomrule
\end{tabular}
}
\end{sc}
\end{small}
\end{center}
\vskip -0.3in
\end{table}


\hypertarget{4.2.2}{\textbf{Transferability in unseen-weather dataset}}: We further evaluate the
transferability of our camouflage under unseen weather conditions.
As shown in Table \ref{sim-table3}, our method achieves the best attack performance on all types of detectors. Our method outperforms the previous state-of-the-art method, ACTIVE, by 23\% on the white-box model YOLOv3, and approximately 11\% on other black-box models. Our camouflage shows better attack performance under new weather conditions than under the weather conditions used for texture generation. This highlights the transferability of our approach across diverse weather conditions. Such good transferability stems from the use of the multi-weather dataset with various fog levels and lighting conditions for texture optimization. The dataset includes 16 different weather conditions, including extreme conditions, such as dark light and dense fog. In the dense fog weather, the vehicle camouflage is heavily obscured but still effective against detectors. The weather in real life is typically not so extreme, so our camouflage can transfer well and achieve effective attacking performance in the unseen conditions. 

Figure \ref{exp2} shows an example of different camouflage on rainy days. Notice that rainy weather is excluded from texture generation. Even though the rain blurs part of the texture, RAUCA uniquely succeeds in attacking the target detector.

\subsection{Evaluation in Real-World Settings}

In this section, we move our test to the real world. We conduct experiments with five types of camouflage in multi-light and foggy conditions to demonstrate the robustness of our camouflage in different physical environments.


\begin{table}[t]
\vskip -0.08in
\caption{Attack comparison under diverse real-world environmental conditions. Values are the car AP@0.5 (\%) with YOLOv3.}
 \vskip -0.12in
\label{phy-table1}
\begin{center}
\begin{small}
\begin{sc}
\resizebox{1\columnwidth}{!}{
\begin{tabular}{lccccc}
\toprule
\multirow{2}*{\textbf{Methods}} & \multicolumn{4}{c}{\textbf{Environmental condition}} & \multirow{2}*{Total}\\ 
\cmidrule(lr){2-5}
                         & Noon            & Afternoon             & Night    & Fog   \\ 
\midrule
Normal                   & 0.810              & 0.82         & 0.728           & 0.721   & 0.770\\
 DAS                      & 0.778              & 0.781         & 0.670         & 0.793     & 0.756\\
 FCA                      & 0.488             & 0.547          & 0.462         & 0.473   & 0.493\\
 DTA & 0.382             & 0.438         & 0.379            & $\bm{0.158}$ & 0.339\\ 
ACTIVE & 0.253             & 0.264         & 0.242         & 0.163   & 0.231\\ 
\midrule
RAUCA                     & $\bm{0.252}$              & $\bm{0.191}$         & $\bm{0.137}$ & 0.200     & $\bm{0.195}$ \\ 
\bottomrule

\end{tabular}
}
\end{sc}
\end{small}
\end{center}
\vskip -0.3in
\end{table}

\begin{figure*}[ht]
\vfill
\begin{center}
\centerline{\includegraphics[width=\columnwidth*2]{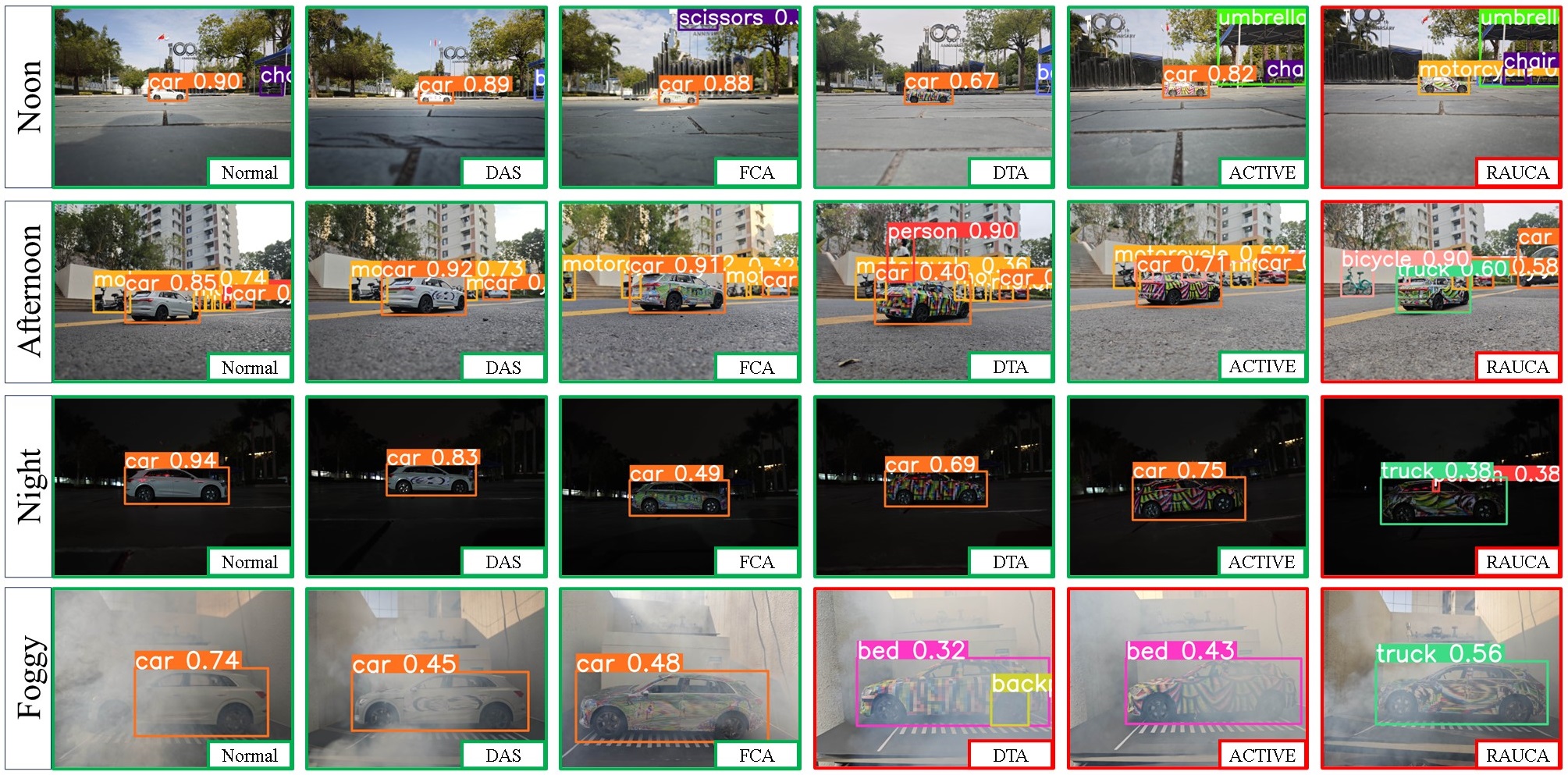}}
\vskip -0.1in
\caption{Real-world evaluation using five different types of camouflage in different environmental conditions.}
\label{real_world}
\end{center}
\vskip -0.4in
\vfill
\end{figure*}

\textbf{Robustness in multi-light conditions}: We change the lighting conditions by taking photos at different times of the day. We report the car AP@0.5 with YOLOv3. The results are shown in Table \ref{phy-table1}, which shows our method achieves the best attack performance in all light conditions. Although our camouflage has a similar attack effectiveness to ACTIVE at noon, it significantly outperforms all other methods, particularly under low-light conditions at night. We can see that almost all of the adversarial camouflage, except DAS, attack better in the real world than in the simulation world, probably because our camera viewpoint chosen in the real world is more susceptible to attacks. The first three rows of Figure \ref{real_world} show the sample prediction results at
different light conditions. Our method distinguishes itself as the only successful approach to deceive the target detector consistently across diverse lighting conditions in these examples.

\textbf{Robustness in foggy conditions}:
We compare the effects of various attack textures in real-world fog scenarios, simulated using a smoke machine. The car AP@0.5 for YOLOv3 detection is reported in the fifth column of Table \ref{phy-table1}. DTA, ACTIVE, and our method demonstrate increased effectiveness in attacking, with DTA and ACTIVE slightly surpassing our method. Given the challenge of maintaining consistent fog density, we acknowledge this discrepancy and consider it reasonable. Figure \ref{real_world} provides examples of vehicle texture detection in foggy environments. It is evident that both DAS and FCA struggle to effectively attack the target detector, while DTA, ACTIVE, and RAUCA successfully induce misclassification in the target detector's output.

We average the detection results for all environmental cases, and the results are in the last column of Table \ref{phy-table1}, which shows that our method has the best attack results. It illustrates the robustness of our method in different physical environments.
\subsection{Ablation Studies}

\begin{table}[t]
\vskip -0.1in
\caption{Comparing the impact \(W (x_{ref})\) on NRP training. Values are Mean Absolute Error values between the output of NRP and ground truth within the car area.}
\label{abl-table0}
\vskip 0.1in
\begin{center}
\begin{small}
\begin{sc}
{
\begin{tabular}{lccccc}
\toprule
\multirow{2}*{\textbf{Methods}} & \multicolumn{4}{c}{\textbf{Distance}}          \\
\cmidrule(lr){2-5}
 & 5           & 10           & 15          & 20           \\
\midrule
Without \(W (x_{ref})\)                   & 7.22              & 6.91                   & 7.14           & 6.76       \\ 
With \(W (x_{ref})\)                  & 6.50              & 5.51                   & 5.53      & 5.23           \\ 

\bottomrule
\end{tabular}
}
\end{sc}
\end{small}
\end{center}
\vskip -0.3in
\end{table}
\textbf{Effectiveness of the \(W (x_{ref})\) for NRP training}: During the training of the NRP, we utilize \(W (x_{ref})\), a weight function that deflates the loss according to the vehicle area in the image, to balance NRP's rendering optimization across various camera viewpoints. In this section, we evaluate the compact of \(W (X_{ref})\) on the accuracy of NRP rendering.
  We report the MAE values between the output of NRP and ground truth within the car area. As shown in Table \ref{abl-table0}, the incorporation of  \(W (x_{ref})\) enhances the rendering capability across all camera distances. Notably, we observe a more pronounced improvement in rendering ability at longer camera distances, aligning with our intended design expectations.

\textbf{Effectiveness of the multi-weather
dataset and NRP for texture generation}: We evaluate the effectiveness of incorporating multi-weather conditions into our dataset and our proposed rendering component, NRP, for texture generation. As can be seen in Table \ref{abl-table1}, we generate the camouflage according to the four combinations ways. The single-weather dataset in the table is identical to the multi-weather dataset, except that the weather parameter is fixed to the default value in Carla. We observe that the multi-weather dataset and the NRP rendering component are crucial to improving attacks\textquotesingle{} performance. Specifically, the NRP rendering component has the most significant impact on adversarial camouflage effects, with an improvement of 0.34\% under the multi-weather dataset.

We also observe that when the framework employs NR as the renderer, introducing multi-weather conditions for texture generation results in a slightly diminished attack performance. This discrepancy may stem from NR's limitations in rendering comprehensive environmental characteristics. The weather information in the multi-weather dataset can't be effectively incorporated into the foreground; instead, it amplifies the contrast between the foreground and background. Consequently, the image obtained from the fusion of the front and rear backgrounds becomes less realistic, causing the texture optimization in the wrong direction.

\textbf{Effectiveness of our adversarial loss for texture generation}: In this section, we evaluate the effectiveness of our designed adversarial loss function. We compare our adversarial loss with the ones used in FCA and ACTIVE. All these losses are calculated using the predicted detection boxes: the object confidence, class confidence, and IOU between the detection box and the ground truth. The difference between our designed loss function and FCA's loss function is that FCA's adversarial loss only considers the output detection box where the ground-truth center is located. In contrast, ours considers all of the detection boxes. It makes the camouflage to attack a broader range of detection boxes, preventing the target from being detected by boxes that do not contain the center of the ground truth. Compared with ACTIVE's, our loss function incorporates the IOU value into the calculation, which makes our camouflage optimization focus more on confusing the detection boxes with a large degree of intersection with the target. Moreover, unlike minimizing the maximum confidence score across all classes in ACTIVE's adversarial loss function, we minimize the maximum confidence score for the car class to achieve a more substantial attack effect. The comparing results are shown in Table \ref{abl-table2}. Values are the car AP@0.5. It can be seen that our proposed adversarial loss function achieves the strongest attack effect on most of the models.

\textbf{Effectiveness of the smooth loss for texture generation}: We investigate the impact of the smooth loss on the attack effectiveness of the generated camouflage. We vary the hyperparameter of the smooth loss $\beta$; the results are shown in Table \ref{abl-table3}. Values are the car AP@0.5. It can be seen that the overall attack effectiveness is better than the previous state-of-the-art ACTIVE method when $\beta$ changes, demonstrating the effectiveness of our generated camouflage. 

\begin{table}[t]
\vskip -0.1in
\caption{The effectiveness of the multi-weather dataset and NRP for texture generation. Values
are car AP@0.5 (\%) with YOLOv3.}
\label{abl-table1}
\vskip 0.1in
\begin{center}
\begin{small}
\begin{sc}
{
\begin{tabular}{lcc}
\toprule
\multirow{2}*{\textbf{Tex-Gen dataset}} & \multicolumn{2}{c}{\textbf{Renderer}}  \\ 
\cmidrule(lr){2-3}
                         & NR            & NRP                  \\ 
\midrule
Single-Weather                 & 0.60             &  0.42                            \\ 
Multi-Weather                   &  0.64          &  $\bm{0.30}$                   \\ 
\bottomrule
\end{tabular}
}
\end{sc}
\end{small}
\end{center}
\vskip -0.1in
\end{table}



\begin{table}[t]
\vskip -0.1in
\caption{The effectiveness of the different adversarial losses for texture generation. Values
are car AP@0.5 (\%) with YOLOv3.}
\label{abl-table2}
\vskip 0.1in
\begin{center}
\begin{small}
\begin{sc}
\resizebox{1\columnwidth}{!}{
\begin{tabular}{lcccccc}
\toprule
\multirow{2}*{\textbf{Methods}} & \multicolumn{3}{c}{\textbf{Single-stage}} & \multicolumn{3}{c}{\textbf{Two-stage}} \\ 
\cmidrule(lr){2-4}\cmidrule(lr){5-7}
                         & YOLOv3            & YOLOX     & DDTR       & DRCN          & SRCN           & FrRCN                   \\ 
\midrule
FCA Loss                   & 0.331              & 0.620         & 0.404     & $\bm{0.437}$           & 0.479     & 0.453             \\ 
ACTIVE Loss                   & 0.348              & 0.625       &   0.344     & 0.439             & 0.462     & 0.424             \\ 
RAUCA Loss              & $\bm{0.304}$       & $\bm{0.611}$   & $\bm{0.285}$     & 0.449        & $\bm{0.384}$     & $\bm{0.406}$             \\
\bottomrule
\end{tabular}
}
\end{sc}
\end{small}
\end{center}
\vskip -0.3in
\end{table}

\begin{table}[t]
\vskip -0.1in
\caption{The effectiveness of the smooth loss for texture generation. Values
are car AP@0.5 (\%) with YOLOv3.}
\label{abl-table3}
\vskip 0.1in
\begin{center}
\begin{small}
\begin{sc}
\resizebox{1\columnwidth}{!}{
\begin{tabular}{lcccccc}
\toprule
\multirow{2}*{\textbf{Methods}} & \multicolumn{3}{c}{\textbf{Single-stage}} & \multicolumn{3}{c}{\textbf{Two-stage}} \\ 
\cmidrule(lr){2-4}\cmidrule(lr){5-7}
                         & YOLOv3            & YOLOX     & DDTR       & DRCN          & SRCN           & FrRCN                   \\ 
\midrule
ACTIVE                   & 0.439              & 0.625         & 0.384     & 0.513             & 0.464     & 0.496             \\
\midrule
$\beta$=0.0                   & 0.343              & 0.619       &   0.307     & 0.454             & 0.419     & 0.431             \\ 
$\beta$=0.0001              & $\bm{0.304}$       & $\bm{0.611}$   & $\bm{0.285}$     & $\bm{0.449}$        & $\bm{0.384}$     & $\bm{0.406}$             \\
$\beta$=0.0003                  & 0.341              & 0.626       &   0.337     & 0.460             & 0.404     & 0.427             \\ 

\bottomrule
\end{tabular}
}
\end{sc}
\end{small}
\end{center}
\vskip -0.3in
\end{table}

\section{Conclusion}
We have proposed RAUCA, a UV-map-based physical adversarial camouflage attack framework with a realistic neural renderer and incorporating multi-weather cases. In particular, we utilize our novel neural render component, namely NRP, which offers the advantages of precise texture mapping and the ability to render environment characteristics. Additionally, we incorporate a multi-weather dataset during camouflage generation to further enhance its robustness. Our experiments demonstrate that RAUCA outperforms the existing works under multi-weather situations, making it more robust in both simulation and physical world settings.

\section*{Acknowledgements}

This work is supported by the National Natural Science Foundation of China (Grants: 62306093, 62376074), the National Key R\&D Program of China (Grant: 2021YFB2700900), the Guangdong Provincial Key Laboratory of Novel Security Intelligence Technologies (Grant: 2022B1212010005), and Shenzhen Science and Technology Program (Grants: JSGGKQTD20221101115655027, KCXST20221021111404010, JSGG20220831103400002, SGDX20230116091244004, KJZD20230923114405011, KJZD20231023095959002, RKX20231110090859012), the Fundamental Research Funds for the Central Universities (Grants: HIT.DZJJ.2023118, HIT.OCEF.2024047),  and the Fok Ying Tung Education Foundation of China (Grant: 171058).

\section*{Impact Statement}

Attackers might exploit our proposed adversarial camouflage method to attack autonomous driving and transportation systems. To remedy this, we suggest enhancing the current vehicle detection models via adversarial training with our camouflage, which can reinforce models against these potential attacks. Additionally, the generated adversarial camouflage can be used to test the models and ensure that the model is robust before deployment. 

\bibliography{example_paper}
\balance
\bibliographystyle{icml2024}

\newpage
\appendix
\onecolumn
\section{Datasets}
This section presents the parameters of all of the datasets in our experiments. 

\textbf{For NRP model training and testing}: We collect vehicle datasets in the CARLA simulation environment under 16 weather conditions. These weather conditions are generated by combining four sun altitude angles (-90\degree, -30\degree, 30\degree, 90\degree) with four fog densities (0, 25, 50, 90). Within each weather
scenario, we randomly choose 30 locations for model training and 3 for testing. For each
car location, imagery is acquired at two azimuth angles
(0\degree, 45\degree), two altitude angles (22.5\degree, 67.5\degree),
and four distances (5m, 10m, 15m, 20m). In terms of color paint for the vehicle,  nine
vehicle colors are utilized during training(red, green,
blue, magenta, yellow, cyan, white, and black and grey ), and 64 vehicle colors (RGB values are taken from
[0, 85, 170, 255] respectively) for testing. In summary, we employ 69,120 photo-realistic images for model training (consisting of 16 weather conditions $\times$ 30 car locations $\times$ 16 camera poses $\times$ 9 colors) and 49152 for testing (consisting of 16 weather conditions $\times$ 3 car locations $\times$ 16 camera poses $\times$ 64 colors)

\textbf{For texture generation and evaluation in physically-based simulation settings}: For the generation of adversarial camouflage, we choose the same 16 weather combinations as NRP training. For evaluation in physically-based simulation, we produce two datasets: one with the same weather combinations as the training set and the other featuring 16 novel weather conditions that are not seen during the texture generation, including diverse scenarios like MidRainyNight and DustStorm. We use the former dataset for robust evaluation and the latter for transferability evaluation. Within each weather
scenario, we randomly choose 20 locations for texture generation and 4 distinct locations for each test dataset. For each car transformation, imagery is captured at every 45-degree increment in azimuth angle, spanning four distinct altitude angles (0.0\degree, 22.5\degree, 45.0\degree, 67.5\degree) and four varying distances (5m, 10m, 15m, 20 m) for each. In summary, we employ 40,960 photo-realistic images for model training (consisting of 16 weather conditions $\times$ 20 car locations $\times$ 128 camera poses $\times$ 1 colors) and 7593 for each test dataset (consisting of 16 weather conditions $\times$ 4 car locations $\times$ 128 camera poses $\times$ 1 colors and obliterating the images of the car entirely obscured by the obstacle).

\textbf{For evaluation in the real world settings}: We print our texture and adversarial texture from DAS \cite{wang2021dual}, FCA \cite{wang2022fca}, DTA \cite{Suryanto_2022_CVPR} as well as ACTIVE \cite{Suryanto_2023_ICCV} and stick them on official 1:12 Audi E-Tron vehicle models. For the adversarial texture generated by the UV-map-based method (DAS, FCA, 
and RAUCA), we print it and paste it on the vehicle directly according to its corresponding position since it is in the form of a UV map. For the world-align-based method (DTN and ACTIVE), since it generates a 2D square texture pattern, it cannot be directly pasted like the UV-map-based camouflage. Hence, we adopt the implementation from previous methods: repeating and enlarging the pattern, printing it, and applying it to the entire vehicle. In addition to this, we also prepare an unpainted vehicle. We shoot these models in the real world with the Xiaomi 14 for comparison.
To demonstrate the robustness of our camouflage in physical multi-light conditions, we capture photos at different times to change the light condition. We take 2592 pictures of six cars with
poses of 3 distances (5cm, 10cm, 15cm), 8 azimuths per 45\degree, and 3
altitude angles (0\degree, 30\degree, 60\degree), as well as at two locations (an asphalt road and an open square) at noon, in the afternoon, and at night
. Moreover, we evaluate textures' effectiveness in the presence
of fog in the real world. We use a smoke machine to simulate the foggy environment on the imitated road on the desktop. We
capture pictures of six cars at every 45-degree increment in azimuth angle, 3 altitude angles (0\degree, 30\degree, 60\degree). We
take five consecutive pictures for the same camera angle with each car
to reduce the unfairness caused by the randomness of the fog. Finally, we obtained 720 pictures for comparison.


\end{document}